# LEBANONUPRISING: A THOROUGH STUDY OF LEBANESE TWEETS

Reda Khalaf[1] and Mireille Makary[2]

[1]Department of Computer Science and Information Technology,
Lebanese International University, Beirut, Lebanon
dev.redakhalaf@gmail.com

[2] Department of Computer Science and Information Technology,
Lebanese International University, Beirut, Lebanon
mireille.makary@liu.edu.lb

## ABSTRACT

*Recent studies showed a huge interest in social networks sentiment analysis such as Twitter, to study how the users feel about a certain topic. In this paper, we conducted a sentiment analysis study for the tweets in spoken Lebanese Arabic related to the LebanonUprising hashtag (#لبنان_ينتفض), which was trending upon a socio-economic revolution that started in October, using different machine learning algorithms. The dataset was manually labelled to measure the precision and recall metrics and to compare between the different algorithms. Furthermore, the work completed in this paper provides two more contributions. The first is related to building a Lebanese – Modern Standard Arabic (فصحة) mapping dictionary and the second is an attempt to detect sarcastic and funny emotions in the tweets using emojis. The results we obtained seem satisfactory especially considering that there was no previous or similar work done involving Lebanese Arabic tweets, to our knowledge.*

## 1. INTRODUCTION

Nowadays, microblogging services such as Facebook and Twitter are considered essential communication tools between people to share their opinions about a certain topic and spread information, and it can all be done in real-time [1]. As published on Statista website, by J. Clement, according to recent social media industry figures, Twitter currently ranks as one of the leading social networks worldwide based on active users. As of the fourth quarter of 2019, Twitter had 152 million monetizable daily active users worldwide. In Lebanon, and as shown by statcounter - GlobalStats[1] for this year, Twitter was mostly used between October and November 2019 and then again between March and April 2020, however Facebook remains the most used social media platform by the Lebanese users. We chose to conduct our analysis using tweets since it was easier to collect the ones related to LebanonUprising topic, while on Facebook, it will be harder to detect the posts, the images that include text about the topic without using any hashtag. But it will definitely be interesting to compare between the two networks in further studies. October 17 was the date when a social-economic revolution started in Lebanon, and users became more active on social media, that could be the interpretation of having the peak of usage of Twitter between October and November. Given the fact that no previous study has been made to tweets in Lebanese Arabic dialect, we decided to conduct a sentiment analysis study of the spoken Lebanese Arabic tweets related to the LebanonUprising hashtag (#لبنان_ينتفض) which was the trending hashtag during the revolution.

---

[1] https://gs.statcounter.com/social-media-stats/all/lebanon

The Arabic language is one of the top five spoken languages in the world [2]. Sentiment Analysis (SA) in Arabic could be a very challenging task since it is rich morphologically and there is always a difference between the formal written Arabic and the daily spoken one [3]. Sentiment analysis also known as opinion mining is a challenging natural language processing or text mining problem [4]. Most research studies treat sentiment analysis as a text classification problem where a particular text is classified as positive, negative or neutral opinion and this process can be automated through the use of machine learning algorithms.

From sentiment analysis, we can study emotions, which are closely related to sentiments, and which are usually subjective feelings and thoughts [5]. According to the study presented in [6], there are six primary emotions shared by people: love, joy, surprise, anger, sadness, and fear, which can be sub-divided into many secondary and tertiary emotions. These emotions can vary in intensity as well. When posting on social media, users frequently use emojis to express their emotions. And therefore, some studies covered the possibility of detecting a particular emotion through certain emojis. We will discuss them in more details in the upcoming sections. The remainder of the paper is divided as follows: the next section covers some of the related work completed in sentiment analysis for Modern Standard Arabic (MSA) tweets and some Jordanian or Saudi dialect tweets. Section 3 describes the experimental design which includes the dataset collected and the tools used, we also go through the manual annotation process for the tweets, the preprocessing steps we applied and then we go through the machine learning algorithms we used to train and predict the sentiments in the tweets. In section 4, we go through the experiments we conducted and compare between the results obtained. We report the accuracy, precision and recall metrics values. We discuss the hypothesis related to automatically detecting sarcastic and funny tweets based on emojis, we report the outcome of the experiments we applied. Since we collected tweets between two different periods, we compared between the users who were active in October and those active between May and August in attempt to study if new users became more involved or if the same users were still using the LebanonUprising hashtag. In the last section, we provide a conclusion of the experiments and we provide a future direction for the work.

## 2. RELATED WORK

SA for Arabic tweets has been an active field for quite some time now, especially considering that it is the native language for 22 countries [7]. Authors in [8] performed opinion mining for tweets targeting unemployment in Saudi Arabia and they faced the challenges related to Saudi dialects compared to MSA, they applied preprocessing techniques to raw data and then used supervised machine learning techniques to analyse sentiments. The classification obtained was satisfying. Another contribution in the field was presented in the Arabic sentiment analysis tool, a lexicon that maps Jordanian Dialect to MSA and a lexicon for emoticons [9]. In their study, the authors collected around 350,000 tweets. Through crowdsourcing, they were able to label more than 25,000 tweets, and then three different machine learning classifiers were used for sentiment analysis. The best accuracy achieved in the experiments the authors reported was obtained using SVM [10] and the score reported was 71.68% when compared to NB [11] and k-nearest neighbours (KNN) [12]. Abdullah et al. [13] performed a comprehensive study on sentiment analysis for Arabic tweets. They reported the challenges in the Arabic domain and not having enough studies that analyse people's opinion in Arabic language. The study demonstrated the need to perform more studies in different Arabic dialects. A hybrid method for sentiment analysis for Arabic tweets for Saudi dialect was also studied in [14]. They provided a two-way classification that classified the tweets as positive or negative, then a three-way classification that led to three classes: positive, negative and neutral and the four-way classification that added the "mixed" class to the three-way classes. The novelty of the work presented in this paper lies in the analysis conducted on tweets based on the spoken Lebanese dialect and in the use of emojis to detect emotions and not just opinion mining.

## 3. EXPERIMENTAL DESIGN

In this section, we discuss the dataset, how it was annotated, the Lebanese – MSA dictionary built, the preprocessing steps and the machine learning classifiers chosen for the experiments.

### 3.1. Dataset

Abed Khooli collected 100K tweets with hashtag LebanonUprising (#لبنان_ينتفض) between October 18th and 21st, 2019 and as mentioned earlier that was following the socio-economic revolution that started on the 17th of October. They were collected in JSON format using workbench data[2], an open source platform for data collection and they were posted on Kaggle[3] which is another platform for sharing code and data related to data science work. Kaggle offers a wide range of public datasets and python notebooks that can be used for data analysis. The tweets in the LebanonUprising dataset were not all in Arabic language, so we had to filter them because our main focus and interest was in the tweets written in spoken Lebanese. Thus, after removing the duplicates, due to the retweets, cleaning the tweets, mostly those consisting of one word, we were left with 21,529 tweets. This dataset was manually labelled, and we will explain in the next section the platform we built for labelling, and it was used to train and build the machine learning models to predict sentiments. We will refer to this dataset as TDS (training dataset).

Then we started collecting tweets with the same hashtag starting May 2020, and until 8 August 2020 using workbench data as well and in the same JSON format. The same cleaning process was applied, and the total number of Lebanese tweets was 24,798 tweets, we will refer to this dataset as PDS (prediction dataset).

### 3.2. Manual labelling of the TDS

In order to be able to build the machine learning models that could predict the sentiment for a Lebanese tweet, we had to train the machine learning classifiers. And in any supervised machine learning algorithm, the classifier has to be trained with labelled data so the accuracy can be measured, and parameters can be tuned to obtain the best model possible for the prediction. Since, to our knowledge, we could not find any labelled dataset using the Lebanese dialect, we decided to label the TDS manually. Hence, we built a web application that would allow the user to classify the tweet into one of the below categories:

1. Sarcastic
2. Angry
3. Negative
4. Neutral (none)
5. Funny
6. Positive

To note that the first 3 categories refer to a negative opinion, the last two refer to a positive opinion. But since we needed to test our hypothesis related to detecting sarcastic or funny emotion through emojis, we kept a flag for the tweets that belong to these two categories. Five users participated in the labelling process. Below is a figure that shows what the platform looked like.

---

[2] https://workbenchdata.com/
[3] https://www.kaggle.com/abedkhooli/lebanon-uprising-october-2019-tweets

Figure 1. Manual labelling web application

### 3.3. Lebanese – MSA Mapping

One of the main steps of the preprocessing included mapping words from the spoken Lebanese to MSA. We needed to make sure that the same word in MSA that could be written in different forms in the Lebanese Arabic is being considered the same when training the classifiers. For that reason, we built a dictionary that maps between the two and we tried to cover as many words included in the tweets as possible. We cannot claim that the work we did is fully complete, but it definitely covered most of the words we could find. And this step could be crucial to apply before we move to the stemming and stopwords removal step. A glimpse of the content of the dictionary is shown in Table 1 below.

Table 1. Lebanese – MSA Mapping

| Lebanese word | MSA Word | English meaning |
|---|---|---|
| تحرفوا \ تحرفو\ تحرف \ بتحرف \ يحرفوا \يحرفو | تحريف | twist |
| يأسانة \ يأست \ يئسوا \ يئسوا\ يئسنا \يأسونا \يئست | يأس | depression |
| تعطو \يعطو\ يعطوا \تعطوا | اعطاء | giving |

### 3.4. Preprocessing

The scripts we used were all written in python and therefore we made use of the available packages to preprocess the text such as removing all the links from the tweets, the emojis, the punctuations, the Arabic stopwords, and then stemming each word using Snowball[4] stemmer that finds the stem or the root of the word after it was mapped to the MSA format.

### 3.5. Supervised Machine learning classifiers

We have selected five supervised machine learning (ML) algorithms to train and use for prediction of sentiments. These algorithms are usually used in case of text classification and for the implementation we used the classifiers implemented in scikit-learn[5]. The algorithms we tested were the Naïve Bayes – MultinomialNB[6], the Support Vector Machines - SVM, the K-Nearest Neighbor – KNN, the SGDClassifier implemented in scikit-learn which is a linear classifier with the Stochastic Gradient Descent – SGD [15] training and the Logistic Regression [16].

---

[4] https://pypi.org/project/snowballstemmer/
[5] https://scikit-learn.org/stable/
[6] https://scikit-learn.org/stable/modules/generated/sklearn.naive_bayes.MultinomialNB.html

# 4. EXPERIMENTS AND RESULTS

We considered the problem at hand as a two-way classification problem. The tweets in the training dataset, TDS, were classified as either positive or negative. Therefore, neutral tweets were discarded, and the angry and sarcastic tweets were considered as a part of the negative tweets. Their count is 8,943 tweets. The funny ones were counted as a part of the positive tweets and their total number was 12,586. We can see clearly that users were more positive than negative in the first few days of the revolution.

To start the experiments, we split the TDS into 85% for training and 15% for testing or as a cross validation step. For each of the machine learning algorithms, we used the grid search class in scikit-learn to look for the best combination of parameters that would lead to the best accuracy on both training and testing sets. We report here the best accuracy outcomes of the different classifiers. The results are detailed in Table 2 below.

Table 2. Accuracy for different machine learning classifiers

| Machine Learning Classifier | Test Set Accuracy |
|---|---|
| SVM | 74.11% |
| LogisticRegression | 73.65% |
| SGDClassifier | 73.77% |
| MultinomialNB | 73.40% |
| KNN | 66.93% |

As we can see, the SVM achieved the best accuracy score. The SGDClassifier, LogisticRegression and MutlinomialNB classifiers had very similar accuracy scores on the test set. The KNN had the lowest accuracy value and that could be due to how the algorithms actually work and the combination of parameters that were more suited for the Lebanese dataset.

## 4.1. Precision and Recall

In addition to the accuracy measure, we computed the precision and recall metrics in an attempt to better understand how well each algorithm is performing. We define the precision metric as being the ratio of true positives (TP) over the sum of true and false (FP) positives. The recall metric is the ratio of the true positives over the sum of true positives and false negatives (FN).

Precision = TP/ (TP+FP)

Recall = TP / (TP + FN)

Accordingly, Table 3 shows how each algorithm performed in labelling the tweets as either positive or negative.

Table 3. Precision and recall metrics for different machine learning classifiers

| Machine Learning Classifier | Precision | Recall |
|---|---|---|
| SVM (C=1, gamma=1, kernel='rbf') | 0.759 | 0.822 |
| LogisticRegression (C=1) | 0.752 | 0.825 |
| SGDClassifier (alpha=0.0001, penalty=l2) | 0.756 | 0.820 |
| MutlinomialNB (alpha=1) | **0.771** | 0.781 |
| KNN (neighbors=6, p=2) | 0.676 | **0.843** |

Based on the scores reported above, we can see that the best precision was achieved using the NB classifier with a value of 0.771 while the highest recall value was obtained using the KNN with a score of 0.843. The algorithms which provided a better accuracy than the NB and KNN had a close precision and recall scores with a difference less than 0.019 for the precision and less than 0.062 for the recall. In all cases, the numbers seem satisfactory when compared to the other Arabic SA studies described in the related work section.

### 4.2. From Sentiment to Emotions using Emojis

As mentioned in [5], emotions are related to sentiments, but they often express a more subjective opinion. Users tend sometimes to use emojis to express their emotions. A recent study exploited emojis for sarcasm detection [17]. The authors showed that the usage of Face with tongue out 😜 emoji is the highest among the sarcastic comments. The Face with tears of Joy 😂, Loud crying face, Grinning and Pouting face are the three specific emojis that are most frequently used with non-sarcastic comments.

So, we considered the study to do some statistics related to the use of emojis in the TDS and we found the results detailed in Table 4.

Table 4.  Statistics related to emojis in tweets

| Label | Total Number of tweets | Tweets contain emojis | Angry Emoji | Sad Emoji | Happy Emoji | Other emojis |
|---|---|---|---|---|---|---|
| Angry | 2,341 | 338 | 92 | 56 | 123 | 166 |
| Sarcastic | 1,803 | 460 | 51 | 85 | 296 | 141 |
| Negative | 4,798 | 626 | 108 | 146 | 255 | 286 |
| Funny | 1,456 | 659 | 69 | 75 | 526 | 148 |
| Positive | 11,130 | 3,302 | 556 | 251 | 1,957 | 1,652 |

Looking at the numbers in Table 4, we can see that emojis were not used frequently in the tweets, however, those labelled as "Funny" had the highest percentage of tweets using emojis and that is ~45%. We looked for particular emojis like the Face with tears of Joy, the face with tongue out and winking eye since they seem to be very used according to [17]. We noticed that 138 tweets out of the 460 sarcastic ones contained the face with tears of joy and 243 out of the 659 funny ones contained the same emoji, that was the highest count for emojis in these two classes. Thus, we hypothesized that when the classifiers will predict negative tweets and in case these tweets contained the face with tears of joy, we will label the tweet as sarcastic, in case the tweet was classified as positive, and had the same emoji, it will be labelled as funny. Validating our hypothesis was done on the PDS, by predicting the sentiment and trying to detect sarcastic and funny ones after prediction.

### 4.3 Prediction on PDS

As described in section 3.1, the PDS consists of 24K+ tweets. It is worth noting that this number of tweets was collected over a duration of 4 months, while nearly the same number was collected in just 4 days at the beginning of the revolution. The same preprocessing steps applied to the TDS were applied to the tweets in PDS. We used the models with the best accuracy reported in Table 3 to label the tweets as either positive or negative. Among the ones predicted as negative, we detected the tweets that contained the face with tears of joy, and we labelled them as sarcastic, the ones predicted positive and that contained that same emoji, were considered funny. Table 5 shows the numbers of tweets predicted using each classifier.

Table 5. PDS sentiment prediction and emotion detection

| ML algorithm | Predicted Positive | Predicted Negative | Detected as Sarcastic | Detected as Funny |
|---|---|---|---|---|
| SVM | 9,100 | 15,698 | 627 | 68 |
| LogisticRegression | 7,446 | 17,352 | 644 | 51 |
| SGDClassifier | 8,638 | 16,160 | 634 | 61 |
| MultinomialNB | 9,955 | 14,843 | 629 | 66 |
| KNN | 15,928 | 8,870 | 585 | 110 |

We can see that in almost all the classifiers, except the KNN, the number of tweets predicted negative is higher than the number predicted positive. We can consider an accuracy of ~74% for the SVM, nearly 73% for the LogisticRegression, SGDClassifier and MultinomialNB and ~67% for the KNN, with a precision greater than 70% for all and a recall greater than 78% as measured on the test set in the TDS.

We did perform a manual verification for the tweets detected as sarcastic or funny. The number of true positive in each case is listed in Table 6.

Table 6. Sarcastic and Funny tweets Validation

| | Sarcastic | | | Funny | | |
|---|---|---|---|---|---|---|
| ML Algorithm | Predicted | True Positive - TP | Accuracy | Predicted | TP | Accuracy |
| SVM | 627 | 404 | 64.43% | 68 | 53 | 77.94% |
| LogisticRegression | 644 | 412 | 63.97% | 51 | 44 | 68.62% |
| SGDClassifier | 634 | 408 | 64.35% | 61 | 51 | 83.60% |
| MultinomialNB | 629 | 405 | 64.38% | 66 | 52 | 78.78% |
| KNN | 585 | 368 | 62.90% | 110 | 57 | 51.81% |

The accuracy in the funny tweets seems better than the one computed for the sarcastic ones because their number is much less than the sarcastic ones. The overall accuracy seems acceptable considering that we are only building our assumption on the existence of one particular emoji in the tweet. It might not be enough for other cases, but it could be something to build on for future work.

### 4.4. Variation in users

The last factor we studied was the number of users who tweeted and re-tweeted in both periods October and between May and August. So, we compared the users in both TDS and PDS, once including the re-tweets and after removing the re-tweets. Therefore, the numbers are as shown in Table 7 below.

Table 7. Comparison between the number of users in TDS and PDS

|  | **Including retweets** | **Excluding retweets** |
|---|---|---|
| **Nb. Of users in TDS** | 38,322 | 9,316 |
| **Nb. Of users in PDS** | 20,850 | 5,406 |
| **Common users between TDS and PDS** | 3,425 | 617 |

From the reported numbers in Table 7, we can see that the number of users tweeting and using the hashtag Lebanon Uprising is reduced, that could indicate a change of interest in the topic for the users. Only 5% of the users tweeted in October and kept tweeting between May and August using the same hashtag while new users started using LebanonUprising hashtag sometime between May and August.

## 5. CONCLUSIONS AND FUTURE WORK

In this paper, we showed the detailed experiments used to build a Lebanese tweets dataset using the LebanonUprising hashtag, how we manually labelled the tweets in the dataset and built a Lebanese – MSA mapping dictionary. This dataset was used to train a set of supervised machine learning classifiers which were then used to predict the sentiments of Lebanese tweets collected over a different period of time. We showed that the performance of the different classifiers was very similar. We have also proposed a hypothesis relating emojis to emotions and in particular sarcasm and funny emotions. We tested our hypothesis on the newly collected tweets, and we obtained satisfactory results. We then compared between the number of users on Twitter tweeting and using LebanonUprising hashtag in October and then between May and August. We showed that new users were tweeting starting May and that only 5% of the users were common between the two timeframes. As a future direction of the work, one can expand the classification from simply opinion mining to emotions detection using not just emojis, but also benefiting from deep neural networks (DNN) which are known to provide satisfactory results in such tasks and words embedding to relate the semantics of the tweets rather than simply considering them as bag of words.

## REFERENCES

[1] Kouloumpis, Efthymios., Wilson, T., Moore, J. (2011) *Twitter sentiment analysis: the good the bad and the OMG!* ICWSM 11, 538–541

[2] Ibrahim, H.S., Abdou, S.M., Gheith, M. (2015) *Sentiment analysis for modern standard Arabic and colloquial.* arXiv preprint, arXiv:1505.03105

[3] Haifa, Kfir & Azmi, Aqil. (2015). *Arabic tweets sentiment analysis - A hybrid scheme.* Journal of Information Science. 42. 10.1177/0165551515610513.

[4] Liu, Bing (2010) *Sentiment Analysis and Subjectivity.* Handbook of Natural Language Processing, Second Edition, (editors: N. Indurkhya and F. J. Damerau)

[5] Wiesław, Wolny (2016) *Emotion analysis of twitter data that use emoticons and emoji ideogram* 25th International Conference On Information Systems Development (ISD2016 Katowice)

[6] Parrott, W (2001). *Emotions in social psychology: Essential readings*. Psychology Press.

[7] Korayem, M., Crandall, D., Abdul-Mageed, M (2012) *Subjectivity and sentiment analysis of Arabic: a survey.* In: Hassanien, A.E., Salem, A.-B.M., Ramadan, R., Kim, T. (eds.) AMLTA 2012. CCIS, vol. 322, pp. 128–139. Springer, Heidelberg. doi:10.1007/978-3-642-35326-0 14

[8] Alwakid, Ghadah, Osman, Taha, Hughes-Roberts, Thomas (2017) *Challenges in Sentiment Analysis for Arabic Social Networks* 3rd International Conference on Arabic Computational Linguistics, ACLing 2017, Dubai, UAE

[9] Duwairi, Rehab & Marji, Raed & Sha'ban, Narmeen & Rushaidat, Sally. (2014). *Sentiment Analysis in Arabic tweets.* 2014 5th International Conference on Information and Communication Systems, ICICS 2014. 1-6. 10.1109/IACS.2014.6841964.

[10] Boser, Bernhard E., Guyon Isabelle M., & Vapnik, Vladimir N. (1992*). A training algorithm for optimal margin classifiers.* In Proceedings of the fifth annual workshop on Computational learning theory (COLT '92). Association for Computing Machinery, New York, NY, USA, 144–152. DOI:https://doi.org/10.1145/130385.130401

[11] Lewis, David. (1998) *Naive Bayes at forty: the independence assumption in information retrieval*. In Machine Learning: ECML-98, Proceedings of the 10th European Conference on Machine Learning, Chemnitz, Germany (pp. 4–15). Berlin: Springer.

[12] Fix, Evelyn, and J. L. Hodges. (1989) *Discriminatory Analysis. Nonparametric Discrimination: Consistency Properties.* International Statistical Review / Revue Internationale De Statistique 57, no. 3: 238-47. Accessed August 26, 2020. doi:10.2307/1403797.

[13] Abdullah, Malak & Hadzikadic, Mirsad. (2017). *Sentiment Analysis on Arabic Tweets: Challenges to Dissecting the Language*. 191-202. 10.1007/978-3-319-58562-8_15.

[14] Al-Twairesh, Nora, Hend Al-Khalifa, AbdulMalik Alsalman, and Yousef Al-Ohali. (2018) *Sentiment analysis of arabic tweets: Feature engineering and a hybrid approach.* arXiv preprint arXiv:1805.08533.

[15] Ruder, Sebastian. (2016). *An overview of gradient descent optimization algorithms.* ArXiv Preprint ArXiv:1609.04747.

[16] LogisticRegression https://scikit-learn.org/stable/modules/linear_model.html#logistic-regression

[17] Subramanian, Jayashree & Sridharan, Varun & Shu, Kai & Liu, Huan. (2019). *Exploiting Emojis for Sarcasm Detection.*

**Authors**

Reda Khalaf: Master student in the Computer Science and Information Technology Department at the Lebanese International University

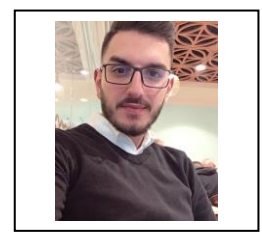

Mireille Makary, PhD: Lecturer in the Computer Science and Information Technology Department at the Lebanese International University with research focus on machine learning and information retrieval.

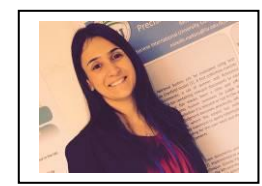